%% file: sample-sigconf.tex
\documentclass[sigconf,natbib=true,anonymous=false]{acmart}

\AtBeginDocument{%
  }

\copyrightyear{2026}
\acmYear{2026}
\setcopyright{cc}
\setcctype{by}
\acmConference[SIGIR '26]{Proceedings of the 49th International ACM SIGIR Conference on Research and Development in Information Retrieval}{July 20--24, 2026}{Melbourne, VIC, Australia}
\acmBooktitle{Proceedings of the 49th International ACM SIGIR Conference on Research and Development in Information Retrieval (SIGIR '26), July 20--24, 2026, Melbourne, VIC, Australia}
\acmDOI{10.1145/3805712.3809693}
\acmISBN{979-8-4007-2599-9/2026/07}

\usepackage{enumitem}
\usepackage{svg}
\usepackage{newunicodechar}
\newunicodechar{♫}{\textmusicalnote}

\begin{document}

% \title{Query-Specific Graph Neural Network for Multi-hop Retrieval Augmented Generation}
\title{Question-Adaptive Graph Learning for Multi-hop Retrieval Augmented Generation}

\author{Yuchen Yan}
\orcid{0009-0005-8672-4468}
\affiliation{%
  \institution{Department of Computer Science, National University of Singapore}
  \country{Singapore}
}
\email{yan1998@comp.nus.edu.sg}

\author{Peiyan Zhang}
\orcid{0000-0002-8691-1846}
\affiliation{
    \institution{Hong Kong University of \\ Science and Technology}
    \country{Hong Kong}}
\email{pzhangao@cse.ust.hk}

\author{Zhihua Liu}
\orcid{0009-0001-7330-5601}
\affiliation{
    \institution{Samsung R\&D institute China-Beijing}
    \city{Beijing}\country{China}}
\email{zhihua.liu@samsung.com}

\author{Hao Wang}
\orcid{0009-0002-5150-5204}
\affiliation{
    \institution{Samsung R\&D Institute China-Beijing}
    \city{Beijing}\country{China}}
\email{hao1.wang@samsung.com}

\author{Yatao Bian}
\orcid{0000-0002-2368-4084}
\affiliation{%
  \institution{Department of Computer Science, National University of Singapore}
  \country{Singapore}
}
\email{ybian@nus.edu.sg}

\author{Weiming Li}
\orcid{0000-0003-4054-5956}
\affiliation{
    \institution{Samsung R\&D institute China-Beijing}
    \city{Beijing}\country{China}}
\email{weiming.li@samsung.com}

\author{Xiaoshuai Hao}
\orcid{0009-0007-4209-6695}
\authornote{Corresponding author}
\affiliation{
    \institution{Xiaomi EV}
    \city{Beijing}
    \country{China}}
\email{haoxiaoshuai@xiaomi.com}

\renewcommand{\shortauthors}{Yuchen Yan et al.}

\begin{abstract}
Retrieval-augmented generation (RAG) has demonstrated its ability to enhance Large Language Models (LLMs) by integrating external knowledge sources. However, multi-hop questions, which require the identification of multiple knowledge targets to form a synthesized answer, raise new challenges for RAG systems. Under the multi-hop settings, existing methods often struggle to fully understand the questions with complex semantic structures and are susceptible to irrelevant noise during the retrieval of multiple information targets. To address these limitations, we propose a novel graph representation learning framework for multi-hop question retrieval. 
We first introduce a Multi-information Level Knowledge Graph (Multi-L KG) to model various information levels for a more comprehensive understanding of multi-hop questions. 
Based on this, we design a Question-Adaptive Graph Neural Network (Quest-GNN) for representation learning on the Multi-L KG. Quest-GNN employs intra/inter-level message passing mechanisms, and in each message passing the information aggregation is guided by the question, which not only facilitates multi-granular information aggregation but also significantly reduces the impact of noise. 
To enhance its ability to learn robust representations, we further propose two synthesized data generation strategies for pre-training the Quest-GNN. 
Extensive experimental results demonstrate the effectiveness of our framework in multi-hop scenarios, especially in high-hop questions the improvement can reach 33.8\%. The code is available at: https://github.com/Jerry2398/QSGNN.
% The code is available at: https://github.com/Jerry2398/QSGNN.
\end{abstract}



\maketitle

\input{chapters/introduction.tex}
\input{chapters/related_work.tex}
\input{chapters/method.tex}
\input{chapters/experiment.tex}

\input{chapters/conclusion.tex}

\appendix
\input{chapters/appendix.tex}

\clearpage

\bibliographystyle{ACM-Reference-Format}
\bibliography{sigir2026_reference}


\end{document}

%% file: chapters/introduction.tex
\section{Introduction}
Retrieval-Augmented Generation (RAG) has emerged as a powerful paradigm for enhancing the capabilities of Large Language Models (LLMs) ~\citep{peng2024graph,gao2023retrieval,liu2025queries,liu2023retrieval,hao2023uncertainty,hao2025dada++,zhang2023continual}. By retrieving pertinent information from external knowledge sources and integrating it into the generation process, RAG enables LLMs to ground their outputs in factual evidence, significantly reducing the propensity for hallucinations ~\citep{fan2024survey,procko2024graph}.

\begin{figure*}[t]
\centering
\includegraphics[width=\textwidth]{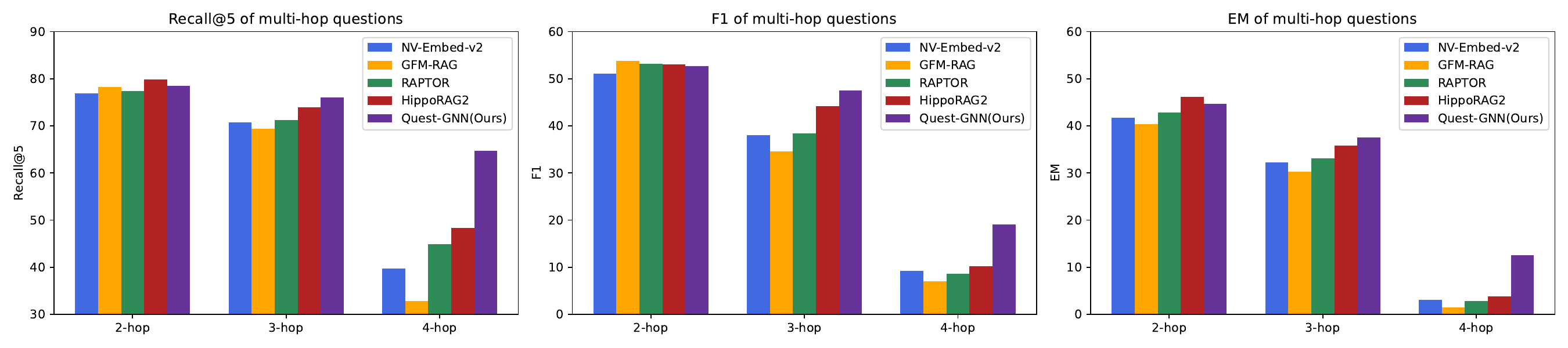}
\caption{
\textbf{Performance on multi-hop questions.}
Existing methods struggle to perform well as hop number increases while Quest-GNN can achieve better performance on high-hop questions.}
\label{fig:multi_hop_res}
\end{figure*}

%%%%%%%%%%% 新改的内容
A challenging scenario in RAG is handling multi-hop questions, where the answer cannot be directly supported by a single document\footnote{Following prior works ~\citep{edge2024local, jimenez2024hipporag, gutierrez2025rag}, we view document as the retrieval unit in this paper.} but requires information synthesized from multiple interrelated documents, each contributing a unique piece of the puzzle for the answer ~\citep{dua2019drop,chen2019multi}. These questions demand that the RAG system accurately identify multiple relevant documents and synthesize them to form a coherent answer. Traditional RAG approaches, which tackle multi-hop questions by decomposing them into a sequence of simpler sub-questions for iterative retrieval~\citep{trivedi2022interleaving,fang2025kirag,wang2025multi,asai2024self,yu2024auto}, often achieve suboptimal performance since they overlook the relational dependencies between documents and may suffer from accumulative intermediate inaccuracies~\cite{hu2024rag,arslan2024survey}. To overcome this limitations, recent efforts have proposed to leverage Knowledge Graphs (KGs). These methods first construct KGs to model the relationships between documents. During retrieval stage, they either navigate the KGs using graph search strategies ~\citep{edge2024local,guo2024lightrag,gutierrez2025rag,chen2025gril} or utilize Graph Neural Networks (GNNs) ~\citep{fang2019hierarchical,mavromatis2024gnn,luo2025gfm,yan2024inductive,zhang2024transgnn,hu2025grag} to identify relevant information. By using KGs, these approaches can better capture the dependencies between documents, thereby improving the retrieval process for multi-hop questions.

% A challenging scenario in RAG is handling multi-hop questions, where the answer cannot be directly supported by a single document\footnote{Following prior works ~\citep{edge2024local, jimenez2024hipporag, gutierrez2025rag}, we view document as the retrieval target in this paper.} but requires information synthesized from multiple interrelated documents, each contributing a unique piece of the puzzle for the answer ~\citep{dua2019drop,chen2019multi}. These questions demand that the RAG system accurately identify multiple relevant documents and synthesize them to form a coherent answer. Traditional RAG approaches, which primarily rely on semantic similarity for retrieval ~\citep{karpukhin2020dense,chen2024bge,jiang2023active,trivedi2022interleaving}, often struggle with multi-hop questions since they overlook the relational dependencies between documents. To overcome this limitation, recent efforts have proposed to leverage Knowledge Graphs (KGs). These methods first construct KGs to model the relationships between documents. During retrieval stage, they either navigate the KGs using graph search strategies ~\citep{edge2024local,guo2024lightrag,gutierrez2025rag} or utilize Graph Neural Networks (GNNs) ~\citep{fang2019hierarchical,mavromatis2024gnn,luo2025gfm} to identify relevant information. By using KGs, these approaches can better capture the dependencies between documents, thereby improving the retrieval process for multi-hop questions.

Although KG based methods have shown improved performance over traditional approaches, they still face significant challenges in addressing the complexities of multi-hop questions. These challenges can be summarized as follows:
i) \textbf{\textit{Semantic Comprehensiveness}}. 
Multi-hop questions are inherently more complex than one-hop questions from the semantic perspective, requiring a comprehensive understanding of diverse information components. 
For instance, consider the question: ``What was Professor Joe's opinion about the Apple event in 2023 that featured the announcement of a foldable smartphone concept?'' In this case, the entity ``Apple'' has multiple possible meanings (\textit{e.g.}, a company or a fruit), and the phrase ``announcement of a foldable smartphone concept'' may refer to a sentence or even a paragraph whose meaning cannot be fully captured by discrete entities within the KG. Such questions not only require the retrieval process to understand context-dependent entities but also demand a full understanding of multi-granular information. However, existing methods primarily focus on entities while ignoring the contextual and multi-granular information. 
This limitation may lead to the inclusion of irrelevant information (\textit{e.g.}, descriptions of apples as fruits) or the loss of critical details (\textit{e.g.}, the absence of precise entities like ``foldable smartphone concept'' in the KG), ultimately compromising the performance.
ii) \textbf{\textit{Noise Sensitivity}}. The retrieval process for multi-hop question is highly sensitive to noise since it has multiple retrieval targets. The noise in any one retrieval target will disrupt the final result. Existing methods are mostly prone to noise. For instance, graph search based methods rely on heuristic strategies to explore relevant information, which is likely to incorporate irrelevant nodes or miss important information. GNN based methods use message passing, which may also aggregate information from irrelevant neighbors. Furthermore, the large search space of KG exacerbates this issue. As shown in Figure ~\ref{fig:multi_hop_res}, the performance of existing methods degrades significantly for high-hop questions, because the number of irrelevant nodes in the KG increases exponentially as the retrieval step increases.

To address these limitations, we propose an effective graph representation learning framework to enhance the RAG performance on multi-hop questions. Our approach is built on three key innovations:
i) To comprehensively capture multiple information granularities and their complex relationships, we propose a \textbf{\textit{Multi-information Level Knowledge Graph (Multi-L KG)}}. It  consists of three node levels, each representing distinct information granularity. The entity-level captures the basic semantic relationships, the chunk-level models the local contextual relationships, and the document-level represents the global thematic relationships. This design encompasses the fine-grained information relationships and facilitates a more comprehensive understanding for multi-hop questions. 
ii) To learn representations from the Multi-L KG for multi-hop retrieval, we design a novel GNN model called \textbf{\textit{Question-Adaptive Graph Neural Network (Quest-GNN)}}. Specifically, it adopts two kinds of message passing to handle the multiple information components of Multi-L KG: intra-level and inter-level message passing. Intra-level message passing focuses on the basic semantic relationships and the logical coherence relationships within each level, while inter-level message passing considers local-global relationships across levels. In each type of message passing, it considers both semantic relationships and alignment with the question. This design integrates multi-level information and significantly reduces the influence of irrelevant nodes, resulting in high-quality representations for multi-hop retrieval.
iii) To further enhance the Quest-GNN, we propose two data generation strategies to produce synthesized QA pairs for pre-training. This process requires no human cost and can be conducted after the construction of Multi-L KG. After pre-training, Quest-GNN can be quickly adapted to downstream tasks. We conduct extensive experiments to evaluate our method, which demonstrate that \textbf{\textit{Quest-GNN achieves state-of-the-art performance compared to existing methods, especially in high hop questions the improvement can reach 33.8\%}}. 

%% file: chapters/related_work.tex
\begin{figure*}[t]
\centering
\includegraphics[width=0.99\textwidth]{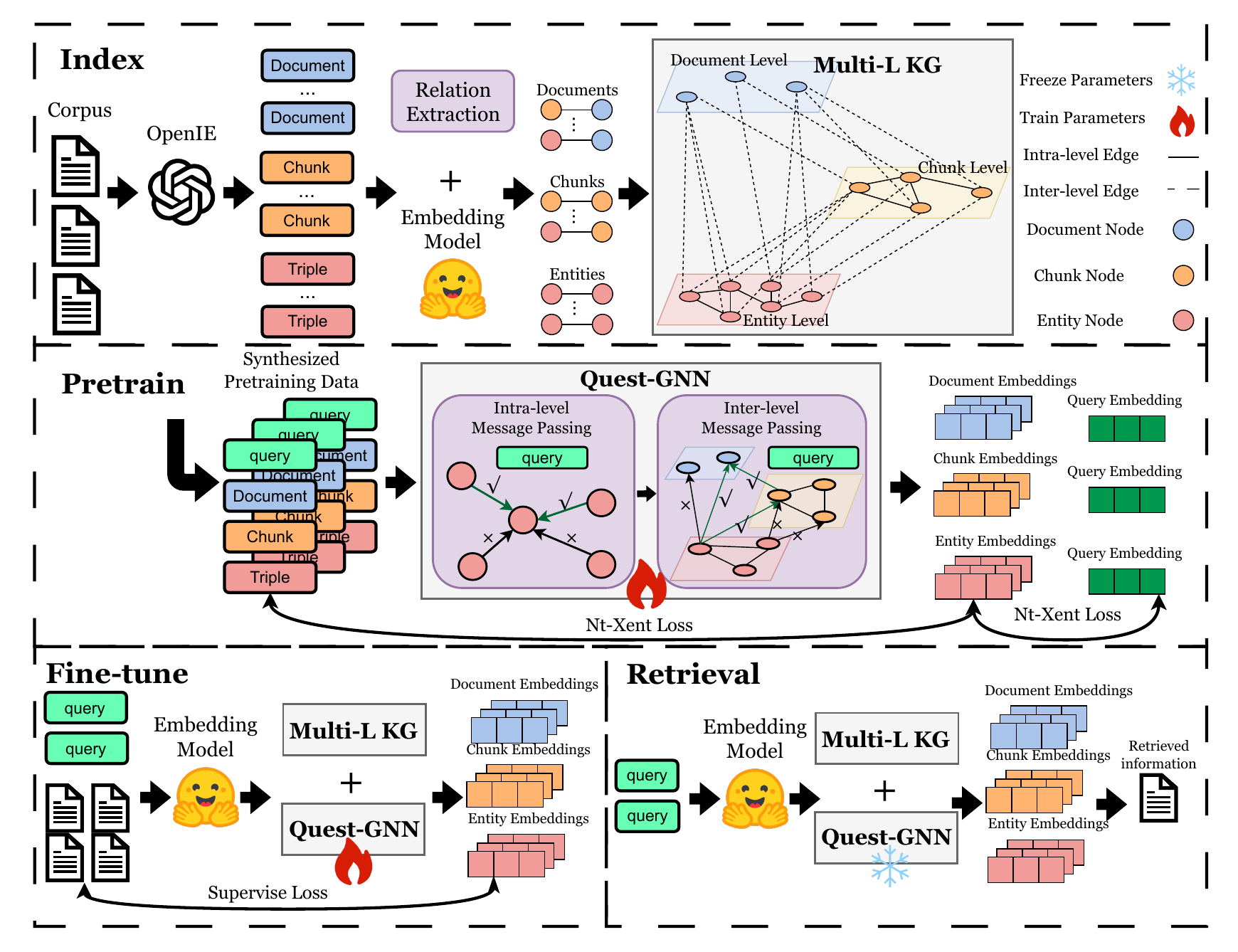}
\caption{\textbf{Framework overview.} It first constructs Multi-L KG to model the multi-level relationships within corpora. Quest-GNN is designed to aggregate information from different levels, all the aggregations are guided by the question. After pre-training and fine-tuning, it can generate representations for multi-hop questions.}
\label{fig:model_framework}
\end{figure*}

\section{Related Work}
\noindent \textbf{Retrieval Augmented Generation (RAG).} RAG aims at enhancing large language models (LLMs) by retrieving relevant information from external knowledge sources and integrating it into the generation process to enable fact-grounded responses ~\citep{gao2023retrieval,fan2024survey,hao2023dual}. 
Text embedding based methods ~\citep{chen2024bge,li2023towards,liu2024unsupervised} encode documents from external knowledge sources into dense embeddings and retrieve relevant information by calculating similarity between question and document embeddings. However, they fail to capture the complex relationships within the questions and the documents and thus struggle to perform well under multi-hop settings. 
In order to adapt the RAG to multi-hop questions, subsequent works ~\citep{jiang2023active,trivedi2022interleaving,su2024dragin,fang2025kirag,wang2025multi,asai2024self,yu2024auto} have proposed multi-step retrieval strategies, where they decompose the original question into multiple sub-questions via LLMs and retrieve relevant information for each sub-question iteratively. 
Although these methods explicitly model the internal structure of the question, they still neglect the multi-level relationships between documents, resulting in limited performance in multi-hop retrievals.

\noindent \textbf{Knowledge Graph for RAG.} Recently, Knowledge Graphs (KGs) have been proposed to facilitate multi-hop retrieval by explicitly modeling complex semantic relationships within and across documents~\citep{peng2024graph,procko2024graph,zhang2024gpt4rec}. Current KG-based Retrieval-Augmented Generation (KG-RAG) can be categorized into two main approaches: Graph search based methods and Graph Neural Network (GNN) based methods. \textbf{Graph search based methods} identify seed nodes and leverage heuristic graph search algorithms (such as breadth-first search, community detection, or LLM reasoning) to traverse the KG and retrieve relevant information for multi-hop questions~\citep{edge2024local, guo2024lightrag, jimenez2024hipporag, gutierrez2025rag,sun2023think,ma2024think,luo2023reasoning}. While these methods are intuitive, they are often susceptible to noise due to their dependence on heuristic strategies, which can introduce irrelevant nodes into the retrieval path. Additionally, the large search space of KGs amplifies the impact of noise, leading to information loss or inaccuracies. In contrast, \textbf{GNN based methods} utilize the message passing mechanism of GNNs to directly output retrieval results from the KG in an end-to-end manner~\citep{luo2025gfm,wang2022vqa} or to first identify seed nodes and then retrieve information using graph search algorithms~\citep{mavromatis2024gnn,fang2019hierarchical,yasunaga2021qa,li2025graph,he2024g}. These methods excel at modeling complex node dependencies through iterative information propagation; however, they also face challenges related to noise, as nodes may aggregate misleading information from irrelevant neighbors. Furthermore, GNN methods often struggle to capture multi-granular semantic information, limiting their effectiveness in addressing complex multi-hop questions.

%% file: chapters/method.tex
%%%%%%%%%%%%%%%%%%%%%%%%%%%%%%%%%%%%
\section{Approach}
This section presents the architecture of our proposed method, whose framework is shown in Figure ~\ref{fig:model_framework}. We first introduce the construction of the \textbf{\textit{Multi-information Level Knowledge Graph (Multi-L KG)}}. Next, we describe the detailed design of the \textbf{\textit{Question-Adaptive Graph Neural Network (Quest-GNN)}}. Finally, we present the pre-training strategy designed to enhance the performance of Quest-GNN in multi-hop question retrieval tasks.

\subsection{Multi-information level KG Construction}
Multi-hop questions are inherently more semantically complex than one-hop questions, as they often encompass not only simple entities but also complex expressions whose meanings may extend beyond the coverage of entities within KGs. Current approaches primarily focus on entities while neglect the multi-granular information, limiting their effectiveness in handling multi-hop questions. To address this limitation, we propose the \textbf{\textit{Multi-information Level Knowledge Graph (Multi-L KG)}}, which integrates various granularities of information to model relationships across different semantic perspectives.

We define the Multi-L KG as $G = \left( \mathcal{O}, \mathcal{C}, \mathcal{D}, \mathcal{E}_{oo}, \mathcal{E}_{oc}, \mathcal{E}_{od}, \mathcal{E}_{cc}, \mathcal{E}_{cd} \right)$. It contains three types of node sets: entity set $\mathcal{O}$, chunk set $\mathcal{C}$, document set $\mathcal{D}$ and five types of edge sets: entity-entity $\mathcal{E}_{oo}$, entity-chunk $\mathcal{E}_{oc}$, entity-document $\mathcal{E}_{od}$, chunk-chunk $\mathcal{E}_{cc}$, chunk-document $\mathcal{E}_{cd}$. The construction of Multi-L KG has two main steps:

\textbf{Node Extraction}. 
Given the document corpus $\mathcal{D}$, we utilize $OpenIE(.)$ ~\citep{angeli2015leveraging,etzioni2008open,zhou2022survey} to extract chunks from each document and derive triples from these chunks\footnote{The prompt for the OpenIE extraction is detailed in the Appendix ~\ref{appendix:prompt_all}}. Note that the term ``chunk'' can have various meanings and in this paper we view sentence as chunk because generally sentence is enough to act as a meaningful reference for questions ~\citep{guo2024lightrag}. The subjects and the objects within the triples are collected as entities. The extraction process can be formalized as:
\begin{equation}
\begin{aligned}
    & OpenIE(\mathcal{D}) \rightarrow \{ (d_{i}, \mathcal{C}_{d_i}, \mathcal{T}_{d_i}) | d_{i} \in \mathcal{D}\} \\
    & \mathcal{C}=\{\mathcal{C}_{d_i}|d_{i} \in \mathcal{D}\}, \mathcal{O}=\{o_{i}|o_{i}\in\mathcal{T}_{d_i},d_{i} \in \mathcal{D}\},
\end{aligned}
\label{method:openie}
\end{equation}
where $d_{i}$ represents the document $i$, $\mathcal{C}_{d_i}$ denotes the set of chunks extracted from $d_{i}$, $\mathcal{T}_{d_i}$ is the set of triples derived from $d_{i}$, $o_{i}$ refers to the subject or object within these triples $\mathcal{T}_{d_i}$, $\mathcal{C}$ denotes the chunk set extracted from $\mathcal{D}$ and $\mathcal{O}$ is the entity set.

\textbf{Relation Construction}.
There are five kinds of relationships ($\mathcal{E}_{oo}$, $\mathcal{E}_{cc}$, $\mathcal{E}_{oc}$, $\mathcal{E}_{od}$, $\mathcal{E}_{cd}$) need to be built. The entity-entity set $\mathcal{E}_{oo}$ is established based on all extracted triples $\{\mathcal{T}_{d_i}|d_{i} \in \mathcal{D}\}$,  capturing fundamental semantic connections. The chunk-chunk set $\mathcal{E}_{cc}$ is constructed by linking adjacent chunks within a document, as their adjacency reflects a kind of logical coherent expression within a document. The last three edge sets entity-chunk $\mathcal{E}_{oc}$, entity-document $\mathcal{E}_{od}$ and chunk-document $\mathcal{E}_{cd}$ are constructed based on containment relationships.

We argue that Multi-L KG provides a comprehensive framework for modeling relationships from multiple perspectives. First, it captures both basic semantic relationships ($\mathcal{E}_{oo}$) and logical coherence ($\mathcal{E}_{cc}$). Second, $\mathcal{E}_{oc}$ and $\mathcal{E}_{od}$ represent local-to-global relationships, enabling precise understanding of entities in different contexts (e.g., ``Apple'' as a fruit or a company). Finally, $\mathcal{E}_{cd}$ serves as a high-level semantic bridge, connecting chunks to their corresponding documents for broader contextual understanding.

\subsection{Question-Adaptive Graph Neural Network}
After constructing the Multi-L KG, the next key challenge is how to effectively retrieve relevant information for multi-hop questions. Existing methods face limitations in their capacity to integrate diverse information levels, and are likely to introduce irrelevant noise due to the complexity of Multi-L KG. To address these challenges, we propose a novel \textbf{\textit{Question-Adaptive Graph Neural Network (Quest-GNN)}}. This approach adaptively aggregates information based on the question, effectively mitigating noise while incorporating insights from all information levels within the Multi-L KG, to generate comprehensive, question-adaptive representations for retrieval. Quest-GNN has two kinds of message passing process: intra-level message passing and inter-level message passing. These two processes are described in detail as follows:

\textbf{Intra-level Message Passing}. Intra-level message passing aggregates information within the same level(i.e. $\mathcal{E}_{oo}$ and $\mathcal{E}_{cc}$). For each information level, the process is defined as:
\begin{equation}
\begin{aligned}
    & \alpha_{i,j} = Sim\left(\mathbf{h}^{l-1}_{i} W^{q}_{\alpha}, \mathbf{h}^{l-1}_{j} W^{k}_{\alpha} \right),\\
    & \beta_{i,j} = Sim\left(\mathbf{q}W^{q}_{\beta}, (\mathbf{h}^{l-1}_{i}||\mathbf{h}^{l-1}_{j})W^{k}_{\beta} \right),\\
    & intra\_attn = Softmax \left( \left\{ \alpha_{i,j} + \beta_{i,j} | j \in \mathcal{N}(i) \right\} \right) , \\
    & msg_{i} = \sum\limits_{j \in \mathcal{N}(i)} intra\_attn_{i,j} \mathbf{h}^{l-1}_{j}W^{v} , \\
    & \mathbf{h}^{l}_{i} = P \left(Norm\left(\mathbf{h}^{l-1}_{i} + msg_{i}\right)\right) , \\
\end{aligned}
\label{method:intra_message_passing}
\end{equation}
where $\mathbf{h}^{l}_{i/j} \in R^{n}$ is the representation of node $i/j$ in layer $l$, initialized with text embeddings from an embedding model \footnote{Embeddings are compressed to dimension $n$ via linear transformations as information bottlenecks.}. $Sim(.)$ calculates the cosine similarity. $\mathbf{q} \in R^{n}$ is the question embedding.$||$ represents the vector concatenation. $W^{q}_{\alpha/\beta} \in R^{n \times n}$, $W^{k}_{\alpha} \in R^{n \times n}$, $W^{k}_{\beta} \in R^{2n \times n}$, $W^{v}\in R^{n \times n}$ are all learnable parameters. $Norm(.)$ is the normalization function, $P(.)$ is the 2-layers MLP with ReLU ~\citep{glorot2011deep}.

Intra-level message passing captures two types of relationships in the Multi-L KG. One is the basic semantic relationship encoded in $\mathcal{E}_{oo}$ and the other is the logical coherence relationship represented by $\mathcal{E}_{cc}$. The aggregation considers not only connectivity, but also semantic similarity $\alpha_{i,j}$ and question alignment $\beta_{i,j}$, which minimizes noise while ensuring question-aware aggregation.

\textbf{Inter-level Message Passing}. Inter-level message passing aggregates information across different levels of Multi-L KG (i.e., $\mathcal{E}_{oc}$, $\mathcal{E}_{od}$ and $\mathcal{E}_{cd}$). For each aggregation, the process is defined as:
\begin{equation}
\begin{aligned}
    & \mathbf{p}_{i,j} = \left((\mathbf{h}^{l-1}_{i})W^{t}||(\mathbf{h}^{l-1}_{j})W^{s}\right),\\
    & \gamma_{i,j} = Sim\left(\mathbf{q}W^{q}_{\gamma}, \mathbf{p}_{i,j} W^{k}_{\gamma} \right),\\
    & inter\_attn = Softmax \left( \left\{ \gamma_{i,j} | j \in \mathcal{N}(i) \right\} \right) , \\
    & \mathbf{msg}_{i} = \sum\limits_{j \in \mathcal{N}(i)} inter\_attn_{i,j} \mathbf{h}^{l-1}_{j}W^{v} , \\
    & \mathbf{h}^{l}_{i} = P \left(Norm\left(\mathbf{h}^{l-1}_{i} + \mathbf{msg}_{i}\right)\right) , \\
\end{aligned}
\label{method:inter_message_passing}
\end{equation}
where $W^{t}$ and $W^{s}$ are used to project heterogeneous nodes into a shared representation space. $W^{q}_{\gamma} \in R^{n \times n}$, $W_{\gamma}^{k}\in R^{2n \times n}$ are learnable parameters. The other notations follow those in Equation~\ref{method:intra_message_passing}.

The inter-level message passing fuses the local, global relationships from hierarchical levels into representations, which contains comprehensive understanding towards the question.

Based on these two processes, Quest-GNN calculates the representations as follows:
\begin{equation}
\begin{aligned}
    & \mathbf{H}^{l}_{intra} = IntraMQ(\mathbf{q}, \mathcal{O}, \mathcal{C}, \mathcal{E}_{oo}, \mathcal{E}_{cc},  \mathbf{H}^{l-1}) , \\
    &  \mathbf{H}^{l} = InterMQ(\mathbf{q}, \mathcal{O}, \mathcal{C}, \mathcal{D}, \mathcal{E}_{oc}, \mathcal{E}_{od},  \mathcal{E}_{cd},  \mathbf{H}^{l}_{intra}) , 
\end{aligned}
\label{method:qsgnn}
\end{equation}
where $IntraMQ(.)$ refers to the intra-level message passing defined in Equation ~\ref{method:intra_message_passing} and $InterMQ(.)$ is the inter-level message passing defined in Equation ~\ref{method:inter_message_passing}.

\subsection{Training Strategy for Quest-GNN}
To enhance the representation learning ability of Quest-GNN, we first pre-train it on synthesized data and then we fine-tune it with human annotations. The pre-training data are the synthesized (question, document) pairs extracted from the same corpora used to construct the Multi-L KG. These pairs contain one-hop questions and two-hop questions, which are all generated by OpenIE \footnote{The prompts can be found in Appendix ~\ref{appendix:prompt_all}}.

\textbf{One-hop Question}. We generate one-hop questions based on the triples extracted from documents. For each triple-document pair $\{(sbj, verb, obj), d\}$, the question is generated in form of $(?, verb, obj)$ or $(sbj, verb, ?)$ where the answer is $sbj$ or $obj$, the support document is $d$ and the $?$ denotes the question target asking the relation with $sbj/obj$.

\textbf{Two-hop Question}. Two-hop questions are generated based on relation chains formed by shared entities across different documents. For instance, if document $d_{i}$ contains the triple $(sbj_{i}, verb_{i}, ent)$ and document $d_{j}$ contains the triple $(ent, verb_{j}, obj_{j})$ where $ent$ is the common entity, we form the chain as $(sbj_{i}, verb_{i}, ent, verb_{j}, obj_{j})$, the corresponding question is generated from the relation chain in the form $(?, verb_{i}, ent, verb_{j}, obj_{j})$ or $(sbj_{i}, verb_{i}, ent, verb_{j}, ?)$ where the question is about the two-hop relationship between $sbj_{i}$ and $obj_{j}$, the support documents are $d_{i}$ and $d_{j}$ and the answer is $sbj_{i}$ or $obj_{j}$.

The Quest-GNN is pre-trained on the synthesized data using the NT-Xent loss ~\citep{chen2020simple}:
\begin{equation}
\begin{aligned}
    & \mathcal{L}_{\text{NT-Xent}} = -\frac{1}{M} \sum_{i=1}^{M} \log \frac{\exp(\text{sim}(\mathbf{q}_i, \mathbf{h}_i) / \tau)}{\sum_{j\in Neg(i)} \mathbf{1}_{[j \neq i]} \exp(\text{sim}(\mathbf{q}_i, \mathbf{h}_j) / \tau)} , \\
\end{aligned}
\label{method:pretrain_loss}
\end{equation}
where $\mathbf{q}_i$ denotes the question embedding, $h_{i}$ is the representation of support document, $M$ is the batch size, $\tau$ is the temperature parameter and we set it as 1.0 in our implementation, $Neg(.)$ denotes the negative sampling and we employ a hard negative sampling strategy ~\citep{schroff2015facenet,xu2022negative} with sampling number set as 30. 

After pre-training, we fine-tune Quest-GNN on the downstream retrieval task. In this stage, the questions are collected by humans with answers and support documents. We fine-tune Quest-GNN with the same loss function as Equation ~\ref{method:pretrain_loss}.

\subsection{Quest-GNN Retrieval and Generation}
Given a question $\mathbf{q}$, the retrieval process of Quest-GNN is denoted as:
\begin{equation}
\begin{aligned}
    & score = Sim(\mathbf{q}, \mathbf{H}^{l}), \\
    & \mathcal{R} = TopK(\mathcal{D},score), \\
\end{aligned}
\label{method:chunk_retrieval_process}
\end{equation}
where $\mathbf{H}^{l}$ represents the representations calculated by Equation \ref{method:qsgnn}. $TopK(.)$ selects the documents with $K$-highest $score$. The retrieval results are feed into LLM as context to generate response:
\begin{equation}
\begin{aligned}
    & response = LLM(question, \mathcal{R}). \\
\end{aligned}
\label{method:chunk_retrieval_process}
\end{equation}

%% file: chapters/experiment.tex
\section{Experimental Results}
\subsection{Experimental Settings}
\subsubsection{Dataset}
We evaluate the effectiveness of our proposed method on three multi-hop QA benchmarks: \textbf{MuSiQue} ~\citep{trivedi2022MuSiQue}, \textbf{2WikiMultiHopQA(2Wiki)} ~\citep{ho2020constructing} and \textbf{HotpotQA} ~\citep{yang2018hotpotqa}. For a fair comparison, we use the same evaluation set (1000 samples for each dataset) as prior works ~\citep{gutierrez2025rag,jimenez2024hipporag,luo2025gfm}. Additionally, we sample extra 1,000 questions from the original dataset for fine-tuning and 225 questions for validation, without overlap with the evaluation set. The basic statistics of these dataset are shown in Table ~\ref{experiment:dataset} and detailed statistics and sampling strategy are presented in Appendix ~\ref{appendix:exp_dataset}.

\begin{table}[h!]
  \centering
  \caption{Dataset statistics.}
  \begin{tabular}{lccc}
    \toprule
                          & MuSiQue  & 2Wiki    & HotpotQA \\
    \midrule
    \#Entity              & 118,021  & 53,153   & 86,147   \\
    \#Chunk               & 57,887   & 23,023   & 39,830   \\
    \#Document            & 15,803   & 7,403    & 9,811    \\
    \bottomrule
  \end{tabular}
  \label{experiment:dataset}
\end{table}

\begin{table*}[h!]
\centering
\caption{\textbf{Retrieval and QA performance using Llama-3.3-70B-Instruct.} Rec@2/Rec@5 denotes Recall@2/Recall@5. No Retrieval means the LLM response without external knowledge. 
We highlight the best results with \textbf{bold} and the second best results with \underline{under line}. 
We do not report the recall of LightRAG or GraphRAG because they do not retrieve documents.}
\begin{tabular}{c|cccc|cccc|cccc}
\toprule
                      & \multicolumn{4}{c|}{MuSiQue}     & \multicolumn{4}{c|}{2Wiki}       & \multicolumn{4}{c}{HotpotQA}  \\
\midrule
Method                & Rec@2       &  Rec@5        & EM        & F1        & Rec@2     &  Rec@5      & EM        & F1      & Rec@2       &  Rec@5        & EM        & F1  \\
\midrule
No Retrieval          & -           & -             & 17.29     & 26.26     & -         & -           & 36.37     & 41.87   & -           & -             & 36.42     & 46.69 \\
BM25                  & 32.76       & 43.62         & 20.38     & 28.04     & 55.32     & 65.31       & 43.15     & 49.49   & 57.36       & 75.82         & 51.26     & 61.44  \\
Contriever+IRCoT      & 34.92       & 46.27         & 24.29     & 31.62     & 46.6      & 57.53       & 38.02     & 40.90   & 56.21       & 74.37         & 49.89     & 60.30  \\
\midrule
GTE-Qwen2-7B          & 51.24       & 67.72         & 33.60     & 40.9      & 66.73     & 77.78       & 54.11     & 59.61   & 77.24       & 90.15         & 58.61     & 70.97  \\ 
NV-Embed-v2           & 50.12       & 68.19         & 31.66     & 39.26     & 67.16     & 78.66       & 53.78     & 59.99   & \textbf{79.27}  & 92.82     & 59.53     & 72.27 \\
\midrule
Auto-RAG              & 41.37       & 61.40         & 27.36     & 34.89     & 55.63     & 74.94       & 45.28     & 55.41   & 65.81       & 79.50         & 51.21     & 60.33 \\
Self-RAG              & 38.42       & 59.17         & 25.83     & 34.75     & 52.92     & 72.45       & 44.76     & 53.27   & 66.36       & 79.27         & 51.45     & 60.02  \\
KiRAG                 & 42.96       & 63.05         & 28.66     & 36.35     & 58.67     & 79.01       & 48.91     & 59.53   & 67.18       & 81.69         & 52.03     & 63.52  \\
\midrule
GNN-RAG               & 34.13       & 53.72         & 28.29     & 34.09     & 58.67     & 69.14       & 38.46     & 49.77   & 67.18       & 81.69         & 52.03     & 63.52  \\
G-retriever           & 47.14       & 66.46         & 30.13     & 38.04     & 62.38     & 78.05       & 46.84     & 58.10   & 74.16       & 89.49         & 54.73     & 65.90  \\
GFM-RAG               & 49.66       & 67.19         & 30.05     & 39.06     & 62.92     & 79.20       & 48.99     & 59.58   & 74.05       & 88.86         & 55.60     & 67.86  \\
\midrule
GraphRAG              & -           & -             & 34.10     & 41.90     & -         & -           & 52.46     & 63.63   & -           & -             & 59.89     & 73.35  \\
LightRAG              & -           & -             &  6.18     & 13.66     & -         & -           &  6.75     & 22.50   & -           & -             & 12.82     & 26.25  \\
RAPTOR                & 47.68       & 69.55         & 32.46     & 40.30     & 66.45     & 82.67       & 48.57     & 62.15   & 77.83       & \underline{93.19} & 60.21 & 73.25  \\    
ToG2                  & 46.16       & 68.04         & 32.75     & 40.28     & 64.86     & 81.22       & 47.74     & 62.14   & 75.14       & 90.18         & 57.99     & 72.53  \\
HippoRAG2             & \underline{50.42}   & \underline{72.29} & \underline{35.16}   & \underline{42.44} & \underline{73.31} & \underline{88.91} & \underline{56.60}     & \underline{66.18} & 78.25             & 92.56           & \textbf{60.92}    & \underline{74.06}  \\
\midrule
Quest-GNN(Ours)           & \textbf{53.88}  & \textbf{75.23} & \textbf{36.65}  & \textbf{44.93}  & \textbf{74.02} & \textbf{89.47}  & \textbf{57.02} & \textbf{66.83}  & \underline{78.42} & \textbf{93.67} & \underline{60.23} & \textbf{74.44}  \\
\bottomrule
\end{tabular}
\label{experiment:recall_and_qa}
\end{table*}

\subsubsection{Baselines} We compare our method (Quest-GNN) with five baseline categories: 
\begin{itemize}[leftmargin=*,noitemsep,topsep=0pt]
    \item \textbf{Naive Retriever}: we use BM25~\citep{robertson1994some} and Contriever~\citep{izacard2021unsupervised} together with IRCoT~\citep{trivedi2022interleaving} as the naive retrievers.
    \item \textbf{Text Embedding}: we use NV-Embed-v2-7B ~\citep{lee2024nv} and GTE-Qwen2-7B-Instruct~\citep{li2023towards} to generate embeddings for retrieval.
    \item \textbf{Iterative Retriever}: we use Self-RAG ~\citep{asai2024self}, Auto-RAG~\citep{yu2024auto} and KiRAG ~\citep{fang2025kirag} as the multi-step RAG baselines.
    \item \textbf{Graph Search}: we use GraphRAG ~\citep{edge2024local}, RAPTOR~\citep{sarthi2024raptor}, LightRAG~\citep{guo2024lightrag}, ToG2~\citep{ma2024think} and HippoRAG2~\citep{gutierrez2025rag} as the graph search baselines.
    \item \textbf{GNN Retriever}: we use GNN-RAG~\citep{mavromatis2024gnn}, G-retriever~\citep{he2024g} and GFM-RAG ~\citep{luo2025gfm} as the GNN retriever baselines.
\end{itemize}
The settings of baselines can be found in Appendix ~\ref{appendix:exp_baseline}.

\subsubsection{Metrics} Follow prior works ~\citep{gutierrez2025rag,jimenez2024hipporag,luo2025gfm}, we evaluate retrieval performance using recall@2 and recall@5 (top-5 retrieved documents) and Question-Answering(QA) performance with exact match (EM) and F1 scores.

\subsubsection{Implementation Details} Our Quest-GNN is implemented with 2 layers, where each layer has one intra-leverl and one inter-level message passing process. Information dimension is set as 128. All the training uses 2 NVIDIA A100(80G) GPUs. In the pre-training stage, we use 95\% samples for training and 5\% samples for checkpoint selection\footnote{The checkpoint is selected based on the retrieval on the 5\% samples using Recall@5.}. The max epoch number is set as 5. We set learning rate as 1e-4 without weight decay. The checkpoint is saved every 2000 steps. In the fine-tuning stage, the max epoch number is set as 3. We set learning rate as 5e-4. The checkpoint is saved every 100 steps and we select checkpoint based on dev set (225 samples) to evaluate the performance. We use NV-Embed-v2-7B ~\citep{lee2024nv} as the embedding model and we use Llama-3.3-70B-Instruct ~\citep{llama3} as the OpenIE model. We retrieve the top-5 documents for LLM(Llama-3.3-70B-Instruct and GPT-4o-mini ~\citep{gpt4omini}) as context for QA task.

\begin{table*}[h!]
\small
\centering
% \vspace{-3pt}
\caption{\textbf{Performance under different hop numbers.} The best result is shown in \textbf{bold}.}
\begin{tabular}{c|ccccccccc|cccccc}
\toprule
                  & \multicolumn{3}{c}{MuSiQue(Rec@5)}            & \multicolumn{3}{c}{MuSiQue(F1)}                 & \multicolumn{3}{c|}{MuSiQue(EM)}             & \multicolumn{2}{c}{2Wiki(Rec@5)}            & \multicolumn{2}{c}{2Wiki(F1)}                 & \multicolumn{2}{c}{2Wiki(EM)}                \\
\midrule
Method            & 2-hop           & 3-hop          & 4-hop          & 2-hop          & 3-hop          & 4-hop          & 2-hop         & 3-hop          & 4-hop    & 2-hop                    & 4-hop          & 2-hop                   & 4-hop          & 2-hop                   & 4-hop         \\
\midrule
NV-Embed-v2       & 76.86           & 70.71          & 39.67          & 51.09          & 37.98          & 9.24           & 41.67         & 32.27          & 3.07     & 81.40                    & 63.52          & 63.21                   & 42.19          & 59.28                   & 23.35          \\
KiRAG             & 75.42           & 64.57          & 26.31          & 49.24          & 33.21          & 6.88           & 38.93         & 28.28          & 1.22     & 84.58    & 48.19      & 64.89       & 29.88       & 55.08   & 14.76  \\
G-retriever     & 79.16     & 67.41     & 29.86     & 54.29     & 31.06     & 6.60      & 41.02     & 29.38         & 1.66      & 82.36     & 54.19     & 63.05     & 30.67     & 52.48     & 15.63 \\
GFM-RAG           & 78.22           & 69.40          & 32.78          & 53.78          & 34.63          & 7.02           & 40.35         & 30.25          & 1.45     & 83.37                    & 56.09          & 64.36                   & 33.11          & 54.44                   & 18.82          \\
GraphRAG          & -               & -              & -              & \textbf{54.99} & 39.91          & 9.79           & 45.18         & 34.38          & 3.22     & -                        & -              & 67.99                   & 39.52          & 57.24                   & 25.97          \\
RAPTOR            & 77.39           & 71.25          & 44.86          & 53.20          & 38.35          & 8.57           & 42.82         & 33.14          & 2.76     & 85.91                    & 64.75          & 66.13                   & 40.14          & 53.71                   & 20.12          \\
ToG2       & 76.82          & 69.71         & 40.81     & 53.85     & 37.54     & 8.23      & 43.18     & 33.66         & 2.43          & 84.50     & 63.09         & 66.23     & 39.51     & 52.63     & 20.69  \\
HippoRAG2         & \textbf{79.89}  & 73.96          & 48.32          & 53.01          & 44.17          & 10.24          & \textbf{46.2} & 35.76          & 3.78     & \textbf{93.15}           & 65.44          & \textbf{70.50}          & 42.26          & \textbf{61.53}          & 29.32          \\
\midrule
Quest-GNN(Ours)       & 78.54           & \textbf{76.02} & \textbf{64.65} & 52.64          & \textbf{47.47} & \textbf{19.03} & 44.63         & \textbf{37.85} & \textbf{12.50}    & 92.10                    & \textbf{74.88}    & 69.72                & \textbf{50.82} & 60.04                   & \textbf{40.28}  \\
\bottomrule
\end{tabular}
\label{experiment:multi_hop_res}
\end{table*}

\subsection{Retrieval and QA Performance} 
Table \ref{experiment:recall_and_qa} shows the retrieval and QA performance of all the methods. We have following observations: 
\begin{itemize}[leftmargin=*,noitemsep,topsep=0pt]
    \item No Retrieval achieves low EM and F1 scores, demonstrating the importance of external knowledge for the LLM's response.
    \item Quest-GNN outperforms all baseline methods, especially compared to other GNN based methods. Recall@5 achieves +11.9\%, +12.9\% and +5.4\% improvements and F1 score gets +15.0\%, +11.7\% and +9.7\% improvements on MuSiQue, 2Wiki and HotpotQA respectively, showing its effectiveness on multi-hop questions.
    \item The classic retrieval algorithms (BM25/Contriever) and text embedding baselines have limited performance as they fail to capture document relationships which is crucial for answering multi-hop questions.
    \item Iterative retrieval based methods show no benefit to multi-hop questions. On the one hand, they only model the question structure but ignore the multi-level relationships between documents. On the other hand, they significantly suffer from intermediate inaccuracies as the suboptimal results may deflect the whole reasoning process.
    \item Although GNN based methods are designed to capture multi-hop relationships, their message passing mechanisms may inadvertently aggregate irrelevant information, which degrades retrieval effectiveness and LLM's response.
    \item Graph search based methods generally perform better than other baselines, but their performance is still constrained by inherent limitations. The heuristic exploration rules often overlook multi-level information and may favor specific path patterns, resulting in data bias. For instance, the performance of RAPTOR varies a lot since it leverages Gaussian Mixture Model to detect clusters in KGs, which is highly biased on data. Notably, LightRAG struggles to perform well on these datasets, which is consistent with other reports ~\citep{gutierrez2025rag,luo2025gfm}. The main reason is that LightRAG is incompatible with multi-hop retrieval workflows as discussed in ~\citep{luo2025gfm}.
\end{itemize}
We report the QA performance using GPT-4o-mini in Appendix ~\ref{appendix:exp_llm_backbone}.

\subsection{Multi-hop Performance}
In order to analyze the detailed performance on multi-hop questions, we evaluate the performance of the competitive baselines on different hop numbers. The results are shown in Table \ref{experiment:multi_hop_res} and we visualize the performance on MuSiQue as Figure ~\ref{fig:multi_hop_res}. We make the following observations:
\begin{itemize}[leftmargin=*,noitemsep,topsep=0pt]
    \item All the methods exhibit performance decline as hop number increases, because of the exponentially growing search space and noise accumulation.
    \item Iterative retrieval based methods underperform in high-hop scenarios due to: i) the negelect of complex relationships between documents, ii) accumulated errors from intermediate results.
    \item Compared to graph search based methods, existing GNN based methods have more performance decrease in high-hop questions because the message passing aggregates the information of irrelevant nodes without question alignment.
    \item As for Quest-GNN, it achieves competitive performance in 2-hop questions compared to other baselines and the improvements become noticeable as the hop number increases. For example, Quest-GNN gets +33.8\%, +85.8\% and +231\% improvements respectively in terms of Recall@5, F1 and EM on 4-hop questions of MuSiQue. This advantage stems from Quest-GNN's question-guided attention mechanism and multi-level message passing, which significantly reduce the impact of noise and fuse the multi-information levels comprehensively. 
\end{itemize}

\subsection{Efficiency} 
We evaluate the efficiency of retrieval across different methods, which is summarized in Table~\ref{experiment:efficiency}. Iterative retrieval based methods (Auto-RAG and KiRAG) show high efficiency because they have no extra cost of indexing or searching on graph. Graph search based methods (HippoRAG2 and RAPTOR) exhibit higher time cost because the graph search strategies are expensive. GNN based methods (Quest-GNN, G-retriever and GFM-RAG) cost less time since message passing is an efficient way to aggregate graph information ~\citep{hamilton2017inductive}. Quest-GNN shows more retrieval time than GFM-RAG and G-retriever due to the complexity of Multi-L KG.

\begin{table}[h!]
\centering
% \vspace{-3pt}
\caption{\textbf{Retrieval efficiency on MuSiQue.} Time means the average retrieval time of each question.}
\begin{tabular}{ccc}
\toprule
Method                                & Recall@5          & Time(s)     \\
\midrule
NV-Embed-v2                           & 68.19             & 0.036       \\
Auto-RAG                              & 61.40             & 0.056       \\
KiRAG                                 & 63.05             & 0.072       \\
G-retriever                           & 66.46             & 0.051       \\
GFM-RAG                               & 67.19             & 0.086       \\
RAPTOR                                & 69.55             & 0.376       \\
HippoRAG2                             & 72.29             & 0.317       \\
Quest-GNN                                 & 75.23             & 0.118       \\
\bottomrule
\end{tabular}
\label{experiment:efficiency}
% \vspace{-2pt}
\end{table}

\begin{table*}[h!]
\centering
\caption{Ablation study. ``chunk'' means performing QA task with chunk. ``w/o entity'', ``w/o chunk'' and ``w/o doc'' mean no entity, chunk or document in Multi-L KG. ``w/o intra'' and ``w/o inter'' mean no intra-level message passing or no inter-level message passing. ``w/o q\_attn'' means no question alignment. ``w/o pt'' means no pre-training. ``w/o ft'' means no fine-tuning.}
\begin{tabular}{cccccccccc}
\toprule
                    & \multicolumn{3}{c}{MuSiQue(Recall@5)} & \multicolumn{3}{c}{MuSiQue(F1)} & \multicolumn{3}{c}{MuSiQue(EM)} \\
Method              & 2-hop       & 3-hop      & 4-hop      & 2-hop     & 3-hop    & 4-hop    & 2-hop     & 3-hop    & 4-hop    \\
% \midrule
% NV-Embed-v2         & 76.86       & 70.71      & 39.67      & 51.09     & 37.98    & 9.24     & 41.67     & 32.27    & 3.07     \\
% HippoRAG2           & 79.89       & 73.96      & 48.32      & 53.01     & 44.17    & 10.24    & 46.20     & 35.76    & 3.78     \\
\midrule
Quest-GNN               & 78.54       & 76.02      & 64.65      & 52.64     & 47.47    & 19.03    & 44.63     & 37.85    & 12.50    \\
Quest-GNN(chunk)         & -           & -          & -          & 40.18     & 37.55    & 13.26    & 30.25     & 29.41    & 8.71     \\
\midrule
Quest-GNN(w/o entity)    & 60.34       & 43.31      & 25.02      & 27.25     & 16.56    & 2.05     & 19.58     & 9.65     & 1.13     \\
Quest-GNN(w/o chunk)     & 77.65       & 74.99      & 62.60      & 51.18     & 45.82    & 16.55    & 42.74     & 35.69    & 9.08     \\
Quest-GNN(w/o doc)       & -           & -          & -          & 36.54     & 32.49    & 8.01     & 26.19     & 26.57    & 5.02     \\
\midrule
Quest-GNN(w/o intra)     & 71.33       & 68.42      & 43.15      & 39.88     & 33.39    & 6.93     & 28.76     & 22.85    & 2.61     \\
Quest-GNN(w/o inter)     & -           & -          & -          & 40.27     & 35.98    & 4.24     & 28.36     & 19.19    & 2.72     \\
Quest-GNN(w/o q\_attn)   & 76.26       & 71.51      & 53.74      & 48.02     & 40.54    & 9.65     & 40.24     & 30.29    & 6.56     \\
\midrule
Quest-GNN(w/o pt)        & 63.66       & 53.93      & 36.83      & 35.64     & 28.14    & 5.15     & 24.62     & 19.28    & 2.53     \\
Quest-GNN(w/o ft)        & 77.19       & 74.69      & 62.57      & 51.08     & 46.42    & 17.12    & 43.58     & 34.23    & 10.48    \\
\bottomrule
\end{tabular}
\label{experiment:ablation}
\end{table*}

\begin{table}[h!]
\small
\centering
\caption{The influence of GNN layer number.}
\begin{tabular}{ccccccc}
\toprule
                        & \multicolumn{3}{c}{MuSiQue(Rec@5)}               & \multicolumn{3}{c}{MuSiQue(F1)}          \\
\midrule
Method                  & 2-hop          & 3-hop          & 4-hop          & 2-hop         & 3-hop          & 4-hop    \\
\midrule
HippoRAG2               & 79.89          & 73.96          & 48.32          & 53.01         & 44.17          & 10.24    \\
\midrule
Quest-GNN(1 layer)          & 73.81          & 69.14          & 55.13          & 46.25         & 41.86          & 11.46    \\
Quest-GNN(2 layers)         & 78.54          & 76.02          & 64.65          & 52.64         & 47.47          & 19.03    \\
Quest-GNN(3 layers)         & 76.35          & 74.55          & 61.66          & 51.34         & 45.24          & 16.45    \\
Quest-GNN(4 layers)         & 62.60          & 58.29          & 35.37          & 38.93         & 29.15          & 4.61     \\
\bottomrule
\end{tabular}
\label{experiment:multi_hop_res_gnn}
\end{table}

\begin{table}[t]
\small
\centering
\caption{The influence of negative sampling number.}
\begin{tabular}{ccccccc}
\toprule
                        & \multicolumn{3}{c}{MuSiQue(Rec@5)}            & \multicolumn{3}{c}{MuSiQue(F1)}        \\
\midrule
Method                   & 2-hop          & 3-hop          & 4-hop          & 2-hop         & 3-hop          & 4-hop     \\
\midrule
HippoRAG2                & 79.89          & 73.96          & 48.32          & 53.01         & 44.17          & 10.24     \\
\midrule
Quest-GNN(Neg\_N=1)          & 68.67          & 60.27          & 50.82          & 43.72         & 38.25          & 5.18      \\
Quest-GNN(Neg\_N=10)         & 77.97          & 76.16          & 62.14          & 51.03         & 46.68          & 17.62     \\
Quest-GNN(Neg\_N=30)         & 78.54          & 76.02          & 64.65          & 52.64         & 47.47          & 19.03     \\
Quest-GNN(Neg\_N=50)         & 77.18          & 76.69          & 63.51          & 51.34         & 46.99          & 18.36     \\
Quest-GNN(Neg\_N=100)        & 77.09          & 76.16          & 63.85          & 52.95         & 46.37          & 18.22     \\
\bottomrule
\end{tabular}
\label{experiment:neg_num_res}
\end{table}

\begin{figure*}[h!]
\centering
\includegraphics[width=\textwidth]{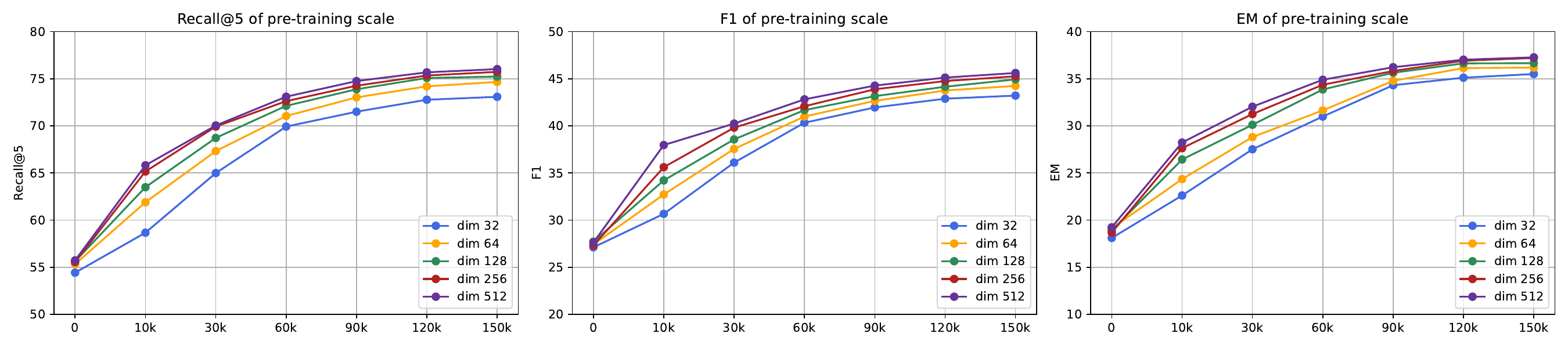}
\caption{
\textbf{Influence of pre-training scale on various model dimension.}}
\label{experiment_fig:scale_vis}
\vspace{-0.5em}
\end{figure*}

\subsection{Ablation Study}
\label{full_paper:ablation}
We present the ablation study of Quest-GNN on Table ~\ref{experiment:ablation}. As for the chunk retrieval (Quest-GNN + chunk, w/o inter or w/o doc), we retrieve the top 10 most relevant chunks for QA and we only report the EM and F1 score since we have no ground truth for chunk retrieval. Our key findings are:
\begin{itemize}[leftmargin=*,noitemsep,topsep=0pt]
    \item Quest-GNN(chunk) does not perform as well as Quest-GNN. It can be attributed to two factors: i) Quest-GNN is not directly trained on chunk labels. ii) Document generally contains more information than chunk.
    \item In terms of the node in Multi-L KG, the model's performance drops significantly without entities, as they form the basic semantic relationships in Multi-L KG. Removing entities causes substantial knowledge loss. Removing chunk nodes shows relatively less impact, as document nodes can provide global contextual information to compensate for their absence. No document lacks global context information, so it does not perform as well as Quest-GNN(chunk). 
    \item In terms of the message passing of Quest-GNN, no intra-level message passing severely deteriorates performance because Quest-GNN loses the ability to directly capture the structure of Multi-L KG but have to be bridged by chunk nodes. No inter-level message passing will not update the representations thus have no difference with the embeddings baselines, whose performance decrease a lot when hop number increases. The absence of question alignment has a particularly negative effect on high-hop questions, where the influence of noise will be amplified as the hop number increases. 
    \item In terms of the training strategy, Quest-GNN(w/o ft) can achieve competitive performance in 2-hop questions but fail to perform as well as Quest-GNN, since we pre-train Quest-GNN only on 1,2-hop questions. Quest-GNN(w/o pt) struggles to perform well because it's likely to overfit with small number of training pairs.
\end{itemize}

\subsection{Hyperparameter Discussion}
We discuss some of important hyper-parameters, including the receptive field (Quest-GNN layer number), negative sampling number, pre-training data scale and information bottleneck dimension.

\subsubsection{Receptive Field Discussion}
We discuss the relationships between GNN layers and the multi-hop performance, the results are shown in Table ~\ref{experiment:multi_hop_res_gnn}. We find that 1-layer Quest-GNN (one intra-level + one inter-level) have limited performance and the gap between 2-layer Quest-GNN becomes bigger as the hop number increase, it is because 1-layer Quest-GNN can only aggregate 2-hop information. The receptive field limits the potential of Quest-GNN. The 2-layer Quest-GNN achieves the best performance among 2,3,4 hop questions because it can aggregate more useful information. However, as the layer number increases (4 layers), the performance will suffer a lot because of the over-smoothing problem of GNN ~\citep{rusch2023survey,chen2020measuring}.

\subsubsection{Negative Sampling Number}
Table~\ref{experiment:neg_num_res} shows the impact of negative sampling number on model's performance. According to the results, we find that low sampling number (Neg\_N = 1) has limited performance due to inadequate discrimination between target documents and distractions. High sampling number (Neg\_N=50, 100) shows limited benefits, because excessive easy negatives fail to provide meaningful guidance for Quest-GNN training.

\subsubsection{Pre-training Scale and Information Dimension}
We pre-train Quest-GNN on the sythesized QA pairs from 0 to 150k across information dimension from 32 to 512 respectively, the results are shown in Figure ~\ref{experiment_fig:scale_vis}. We conduct experiments on the MuSiQue dataset. As for the pre-training scale, the results show that insufficient pre-training data leads to sub-optimal performance across all the dimensions. As the amount of pre-training data increases, the performance of Quest-GNN gets better, although the marginal improvement decreases. As for the information dimension, the results indicate that increasing the information dimension improves the performance of Quest-GNN, as larger dimensions enhance model expressiveness through increased model scale. This observation is consistent with the model scaling analysis presented in GFM-RAG ~\citep{luo2025gfm}. We believe that further improvements can be achieved by combining a larger model with a substantial amount of pre-training data. In this paper, we set the dimension as 128 considering the balance between performance and computational limitation.

%% file: chapters/conclusion.tex
\section{Conclusion}
In this paper, we propose a novel graph learning method for multi-hop questions retrieval. We first design a Multi-information Level Knowledge Graph(Multi-L KG) to model the multi-granular relationships among documents. Then we introduce Question-Adaptive Graph Neural Network(Quest-GNN). The Quest-GNN aggregates information from multi-levels, together with question alignment, to ensure comprehensive question specific representations while filtering out the noise. The pre-training strategy further enhance the ability of Quest-GNN. Extensive experiments shows the effectiveness of our method in multi-hop questions.

%% file: chapters/appendix.tex
\section{Appendix}

%%%%%%%%%%%%%%%%%%%%%%%%
% \appendix
% \section{Appendix}

% This supplementary material provides additional details on the proposed method and experimental results that could not be included in the main manuscript due to page limitations.
% Specifically, this appendix is organized as follows.

% \begin{itemize}
% \item Sec.~\ref{appendix:llm_use} provides the use of Large Language Models (LLMs).
% \item Sec.~\ref{appendix:exp_dataset} provides more details on the dataset settings and data processing.
% \item Sec.~\ref{appendix:exp_baseline} outlines the implementation of baseline models and their experimental settings.
% \item Sec.~\ref{appendix:exp_llm_backbone} presents the QA performance on GPT-4o-mini.
% \item Sec.~\ref{appendix:multi_hop_2wiki} provides the multi-hop performance on 2WikimultihopQA(2Wiki) dataset.
% \item Sec.~\ref{appendix:ablation} provides detailed ablation study for our method.
% \item Sec.~\ref{appendix:hyperparameter_discussion} presents hyperparameter study for our method.
% \item Sec.~\ref{appendix:good_case_study} discusses three good cases of our method compared to HippoRAG2
% \item Sec.~\ref{appendix:bad_case_study} presents three bad cases of our method and discusses the limitation and possible solution as future work.
% \item Sec.~\ref{appendix:openie_prompt} presents the prompts used for OpenIE.
% \item Sec.~\ref{appendix:pretrain_prompt} presents the prompts used for generating pre-training data.
% %shows limitation and future works.
% \end{itemize}

% \label{appendix:sec_content}

%%%%%%%%%%%%%%%%%%%%%%%%
\subsection{Dataset Settings}
\label{appendix:exp_dataset}
\begin{table}[h!]
  \centering
  \caption{Dataset statistics}
  \begin{tabular}{lccc}
    \toprule
                          & MuSiQue   & 2Wiki    & HotpotQA \\
    \midrule
    \#Pre-train 1-hop QA  & 91,621    & 52,122   & 73,128   \\
    \#Pre-train 2-hop QA  & 58,923    & 12,097   & 20,619   \\
    \midrule
    \#Fine-tune 2-hop QA  & 270       & 782      & 1000     \\
    \#Fine-tune 3-hop QA  & 482       & 0        & 0        \\
    \#Fine-tune 4-hop QA  & 248       & 218      & 0        \\
    \midrule
    \#Test 2-hop QA       & 66        & 172      & 225      \\
    \#Test 3-hop QA       & 104       & 0        & 0        \\
    \#Test 4-hop QA       & 55        & 53       & 0        \\
    \midrule
    \#Eval 2-hop QA       & 488       & 847      & 1000     \\
    \#Eval 3-hop QA       & 334       & 0        & 0        \\
    \#Eval 4-hop QA       & 178       & 153      & 0        \\
    \bottomrule
  \end{tabular}
  %\vspace{-15pt}
  \label{appendix:dataset_statistics}
\end{table}

\noindent The statistics of benchmark datasets are shown in Table ~\ref{appendix:dataset_statistics}. For the MuSiQue, 2Wiki and HotpotQA, the extra sample have at least 60\%, 60\% and 50\% supported documents in existing corpora.

\subsection{Baseline Settings}
\label{appendix:exp_baseline}
BM25 is implemented as BM25S~\citep{lu2024bm25s}. As for the text embedding based models, we use the NV-Embed-v2-7B and GTE-Qwen2-7B-Instruct from Huggingface ~\citep{wolf2019huggingface}. For the all the other baselines we use their official implementations. For iterative retrievers, we set the max retrieval number as 5. For GFM-RAG, we use the official settings to pre-train and fine-tune on our corpora. For GNN-RAG we use ReaRev ~\citep{mavromatis2022rearev} as GNN reasoning. All other baseline settings are as their official suggestions.

\subsection{QA Performance on GPT-4o-mini}
\label{appendix:exp_llm_backbone}
%\vspace{-10pt}
\begin{table}[h!]
\small
\centering
\caption{QA performance on GPT-4o-mini. We highlight the best with \textbf{bold} and the second best with \underline{under line}.} 
\begin{tabular}{ccccccc}
\toprule
                      & \multicolumn{2}{c}{MuSiQue}     & \multicolumn{2}{c}{2Wiki}        & \multicolumn{2}{c}{HotpotQA}  \\
\midrule
Method                & EM                  & F1                    & EM                    & F1                    & EM                & F1     \\
\midrule
No Retrieval          & 10.81               & 21.39                 & 31.83                 & 36.51                 & 27.44             & 41.47      \\
BM25                  & 12.54               & 23.88                 & 32.25                 & 40.30                 & 34.50             & 52.62      \\
Contriever+IRCoT      & 13.90               & 25.28                 & 29.44                 & 34.66                 & 30.17             & 45.55      \\
\midrule
GTE-Qwen2-7B          & 31.15               & 39.88                 & 50.75                 & 52.37                 & 52.19             & 66.17      \\
NV-Embed-v2           & 28.07               & 37.15                 & 49.22                 & 51.51                 & 54.85             & 68.43      \\
\midrule
GNN-RAG               & 23.02               & 30.12                 & 34.52                 & 48.25                 & 49.12             & 58.99       \\
GFM-RAG               & 28.30               & 36.81                 & 43.79                 & 53.26                 & 51.86             & 64.73       \\
\midrule
RAPTOR                & 32.93               & \underline{41.04}     & 45.43                 & 56.24                 & 55.19             & 68.16       \\
GraphRAG              & \underline{33.31}   & 40.13                 & 50.99                 & 58.08                 & 56.45             & 69.24       \\
HippoRAG2             & 33.12               & 40.30                 & \underline{51.49}     & \underline{60.18}     & \textbf{56.39}    & \textbf{70.53}  \\
\midrule
Quest-GNN(Ours)           & \textbf{34.27}      & \textbf{41.78}        & \textbf{53.52}        & \textbf{61.63}        & \underline{56.04} & \underline{69.73} \\
\bottomrule
\end{tabular}
%\vspace{-18pt}
\label{appendix:qa_gpt}
\end{table}

Table~\ref{appendix:qa_gpt} presents QA performance using GPT-4o-mini. The results show that Quest-GNN achieves competitive EM and F1 scores, demonstrating its compatibility with different LLM backbones. Notably, all methods show reduced performance compared to Llama-3-70B-Instruct, primarily due to  the intrinsic knowledge gap between GPT-4o-mini and Llama-3.3-70B-Instruct. Other observations align with the report of Table ~\ref{experiment:recall_and_qa}.

\subsection{Case Study and Discussion}
\label{appendix:good_case_study}
We present the case study of Quest-GNN here, which includes two good cases and two bad cases. Correct evidence is marked in green and incorrect evidence in red. Notably, Quest-GNN's marks are based on question alignment, whereas HippoRAG2's marks are based on PPR scores ~\citep{gutierrez2025rag}.

\begin{figure}[h!]
\centering
\includegraphics[width=0.48\textwidth]{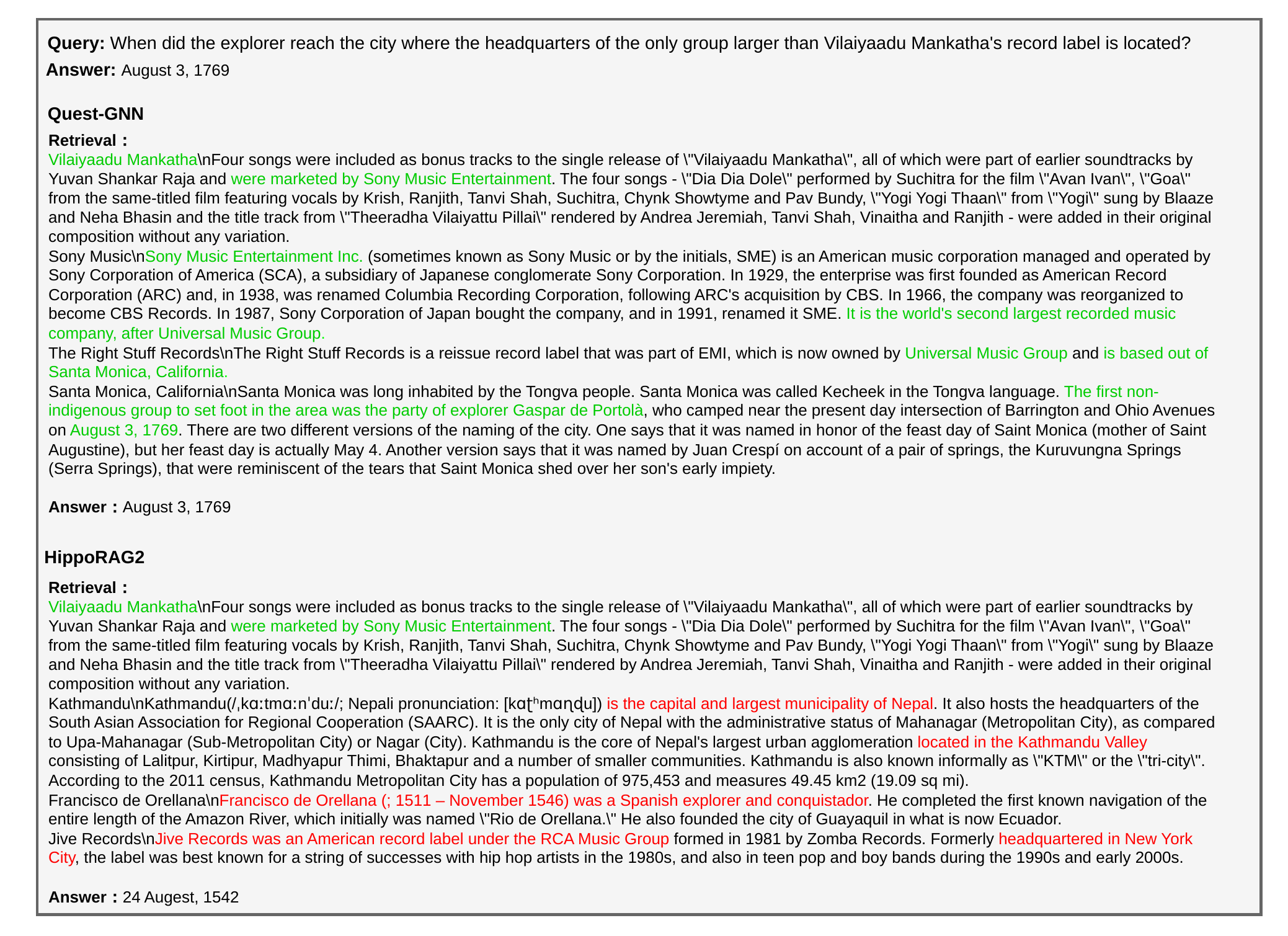}
\caption{
\textbf{Good case 1 for Quest-GNN.}
}
\label{fig:good_case_1}
\end{figure}

\begin{figure}[h!]
\centering
\includegraphics[width=0.48\textwidth]{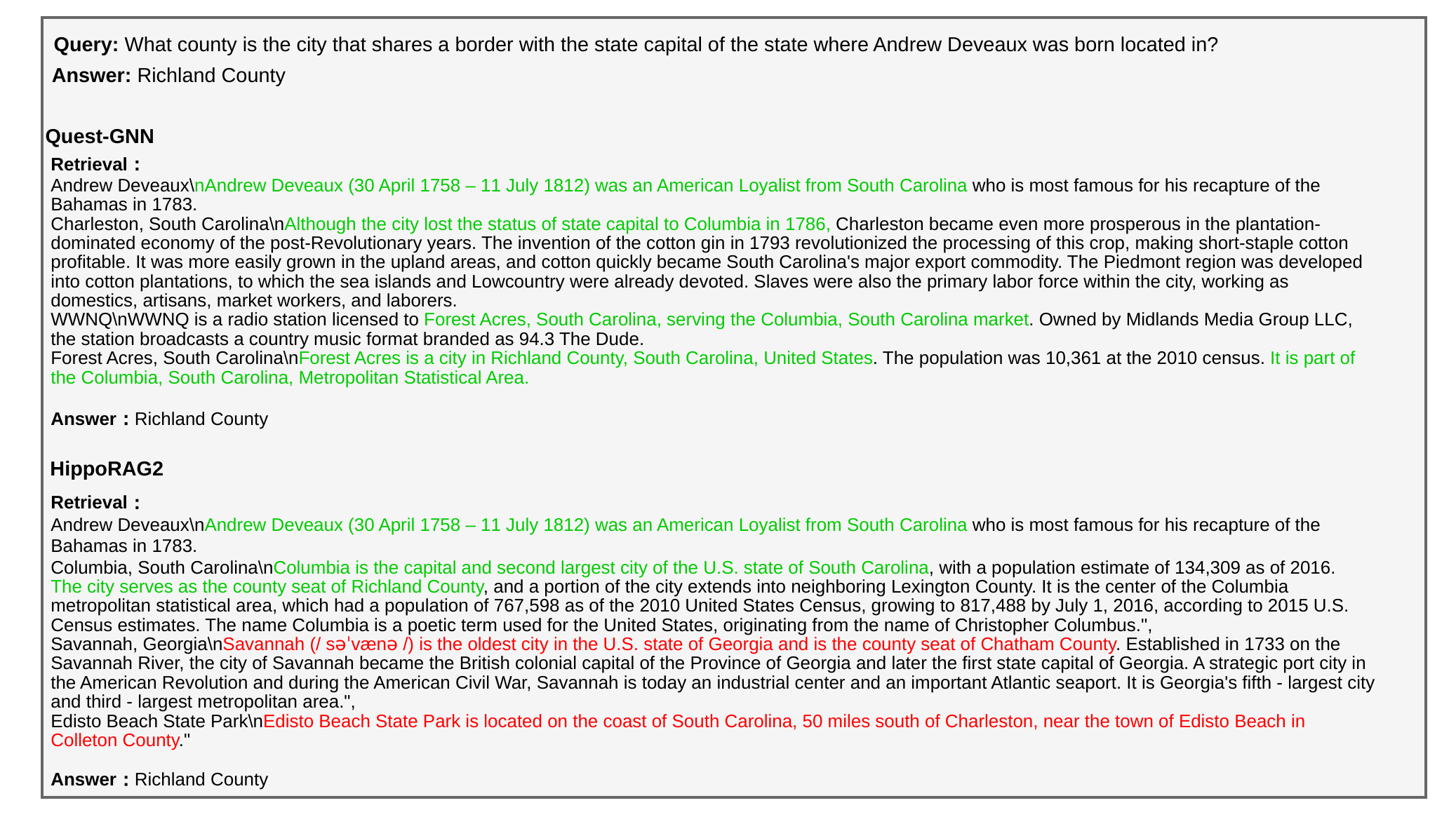}
\caption{
\textbf{Good case 2 for Quest-GNN.}
}
\label{fig:good_case_2}
\end{figure}

\noindent \textbf{Good Case Analysis}. Figure ~\ref{fig:good_case_1} and Figure ~\ref{fig:good_case_2} show two good cases where Quest-GNN achieves recall@5 of 1.0 with correct answer generation while retrieval of HippoRAG2 is inaccurate. Quest-GNN successfully identifies the critical understanding composite semantics (For example, ``the headquarters of'' refers to ``based in Santa Monica, California'' in case 1), and integrates evidence across multiple documents (For example, biographical data:``born in South Carolina'', state capital history:``Columbia became capital in 1786'', and geographic relationships:``Forest Acres borders Columbia'' in case 2). But HippoRAG2 fails to select the right seed nodes without high-level contextual information. These two cases show Multi-L KG's semantic modeling capability and question alignment's effectiveness in filtering misinformation and understanding of contextual information comprehensively.

\begin{figure}[h!]
\centering
\includegraphics[width=0.48\textwidth]{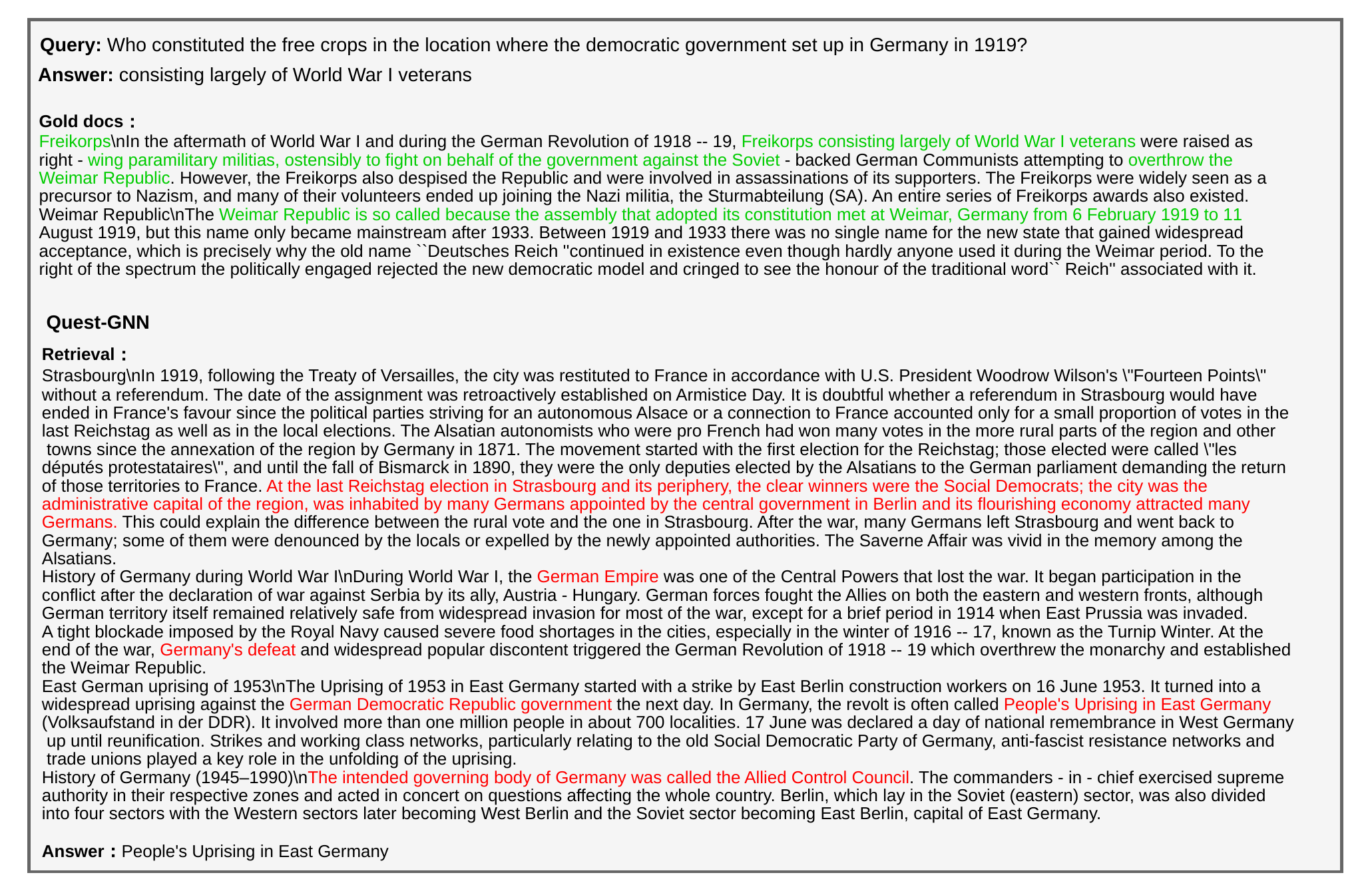}
\caption{
\textbf{Bad case 1 for Quest-GNN.}
}
\label{fig:bad_case_1}
\end{figure}

\begin{figure}[h!]
\centering
\includegraphics[width=0.48\textwidth]{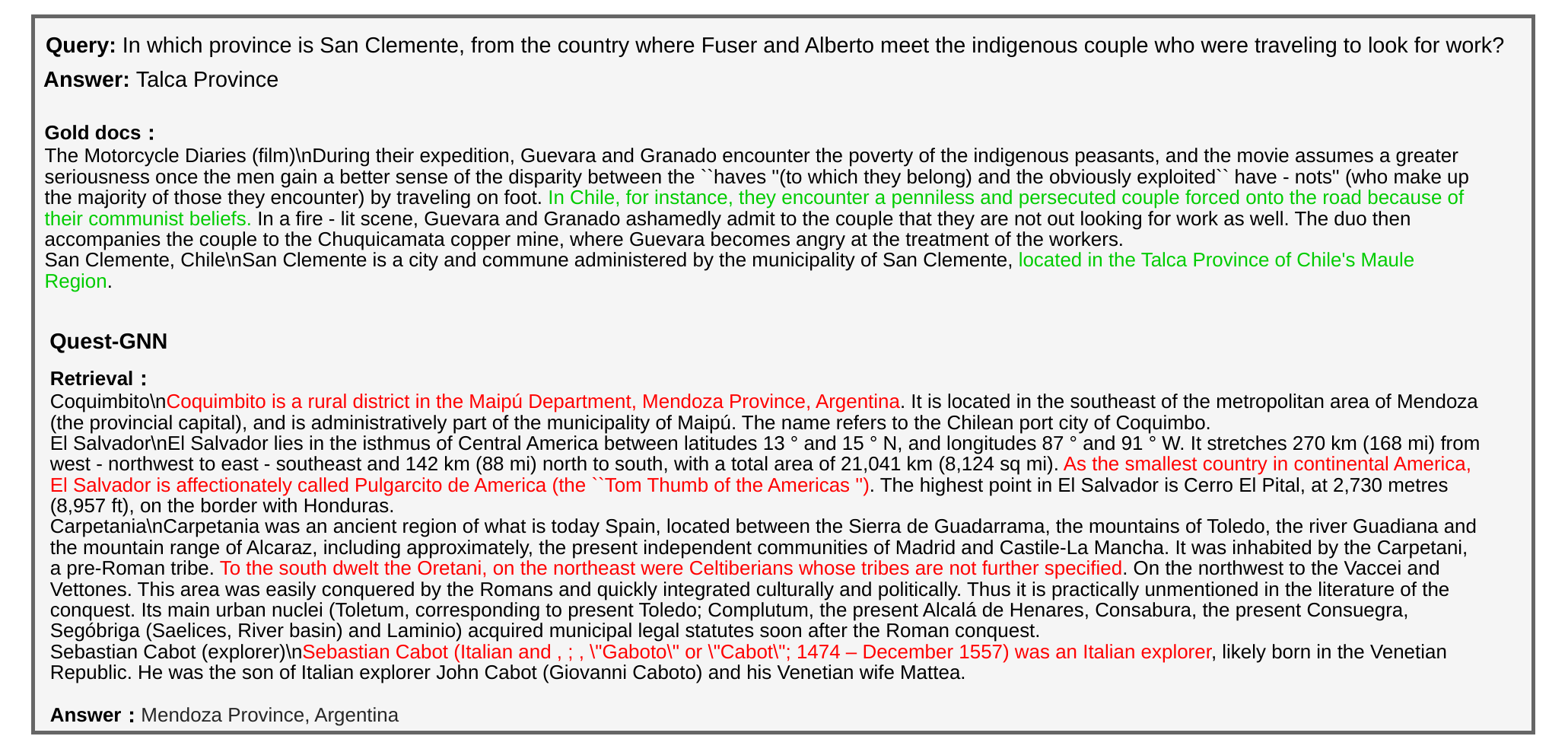}
\caption{
\textbf{Bad case 2 for Quest-GNN.}
}
\label{fig:bad_case_2}
\end{figure}

\noindent \textbf{Bad Case Analysis}. Figure ~\ref{fig:bad_case_1} and Figure ~\ref{fig:bad_case_2} reveal two bad cases for Quest-GNN. These two cases both stem from poor initial representation. In case 1 ``Freikorps'' is a misspelled query terminology, which should be ``free crops'' (we changed to the ``free crops'' then Quest-GNN could retrieve correctly). This leads Quest-GNN to prioritize non-critical entities like ``democratic government'' and ``Germany''. We also check the pre-training data and fine-tuning data, there is no information about the associations between ``free crops'' and ``Freikorps'', which exacerbates the model's confusion. In case 2, ``San Clemente'', ``Fuser'' and ``Alberto'' are all under-represented since they are absent from both pre-training and fine-tuning corpora. However, all of them are important for correct retrieval.

\noindent \textbf{Limitation Discussion and Future Work}. 
Based on the analysis of the bad cases above, Quest-GNN exhibits limitations when dealing with specific terminology or misspelling that can not be well-represented. We think this limitation can be mitigated by selecting seed nodes as domain-specific terminologies and sampling constrained subgraphs to prioritize relevant documents in retrieval. But how to realize this with limited information loss is still non-trivial. We will leave this as our future work.

%%%%%%%%%%%%%%%% prompt
\subsection{Prompts}
\label{appendix:prompt_all}

\begin{figure}[h!]
%\vspace{-10pt}
\centering
\includegraphics[width=0.48\textwidth]{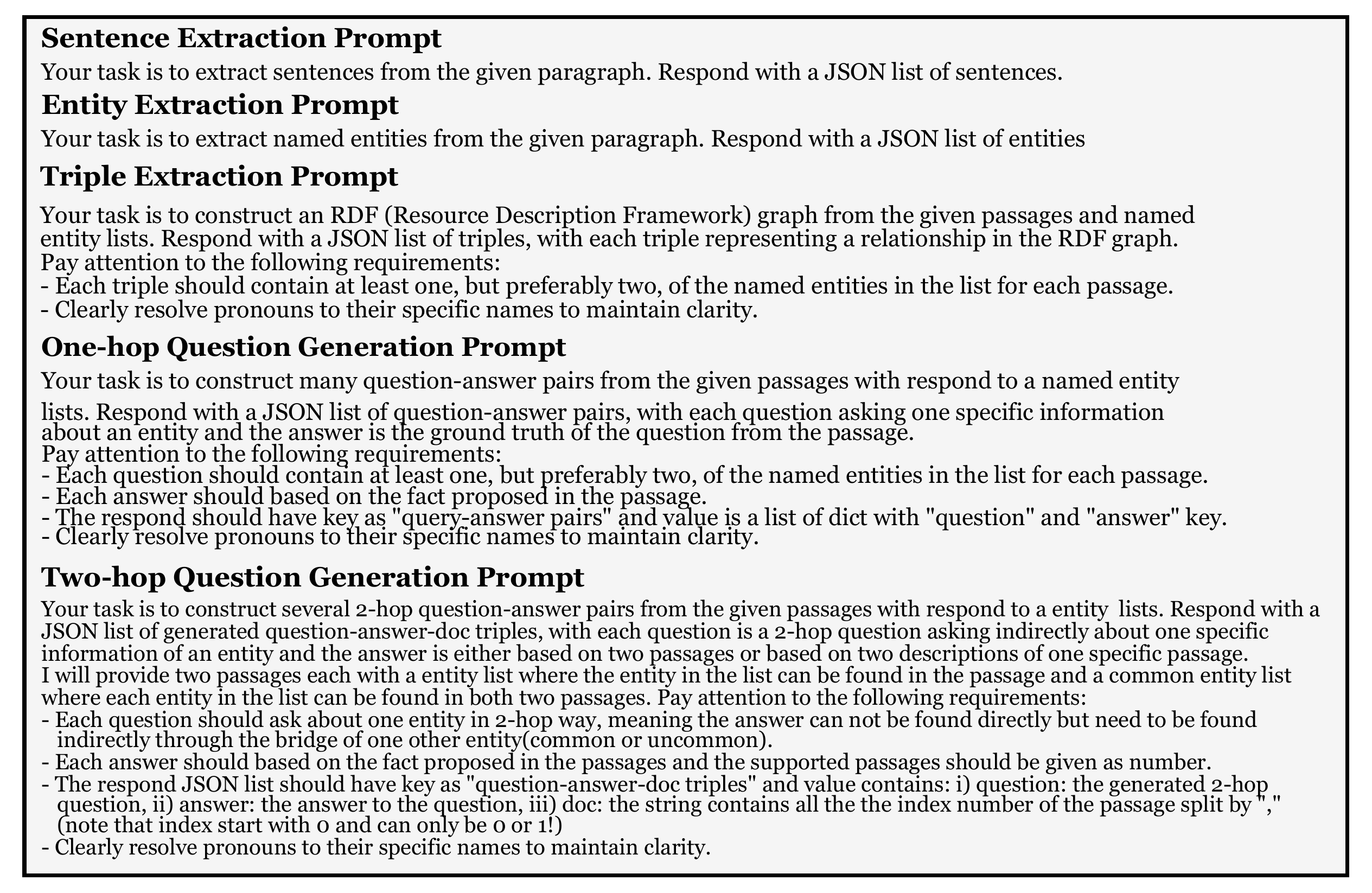}
%\vspace{-10pt}
\caption{
\textbf{Extraction and question generation prompts.}
}
%\vspace{-10pt}
\label{fig:prompt_all}
\end{figure}

The prompts for OpenIE and question generation are in Figure ~\ref{fig:prompt_all}.

